\documentclass{article}

\PassOptionsToPackage{numbers, sort}{natbib}

\usepackage{arxiv}
\usepackage{amsmath}
\usepackage{amsthm}
\usepackage{multirow}

\usepackage{colortbl}

\usepackage[utf8]{inputenc} 
\usepackage[T1]{fontenc}    
\usepackage{url}            
\usepackage{booktabs}       
\usepackage{amsfonts}       
\usepackage{nicefrac}       
\usepackage{algorithm}
\usepackage{algorithmic}

\usepackage{nccmath}

\DeclareMathOperator*{\argmin}{arg\,min}

\usepackage{xcolor}         
\definecolor{citecolor}{HTML}{0071BC}
\definecolor{linkcolor}{HTML}{D32F2F}
\definecolor{cellcolor}{HTML}{E3F2FD}
\definecolor{red}{HTML}{D32F2F}
\definecolor{magenta}{HTML}{D81B60}
\usepackage{hyperref}
\usepackage{url}
\hypersetup{colorlinks=true, linkcolor=linkcolor, citecolor=citecolor,urlcolor=magenta}

\title{ATLAS: Adapter-Based Multi-Modal Continual Learning\\ with a Two-Stage Learning Strategy}
\author{%
    \textbf{
    Hong Li$^{1, 4, }$\thanks{Equal contribution. This work was done when the first author was interned at Shanghai Jiao Tong University.}
    \quad
    Zhiquan Tan$^{2, *}$
    \quad
    Xingyu Li$^{3, *}$
    \quad
    Weiran Huang$^{1,}$\thanks{Correspondence to Weiran Huang (weiran.huang@outlook.com).}}
    \\[0.3cm]
    $^1$ MIFA Lab, Qing Yuan Research Institute, SEIEE, Shanghai Jiao Tong University\\
    $^2$ Department of Mathematical Sciences, Tsinghua University\\
    $^3$ Lin Gang Laboratory \\
    $^4$ ShanghaiTech University
}

\begin{document}

\maketitle

\begin{abstract}
While vision-and-language models significantly advance in many fields, the challenge of continual learning is unsolved. Parameter-efficient modules like adapters and prompts present a promising way to alleviate catastrophic forgetting. However, existing works usually learn individual adapters for each task, which may result in redundant knowledge among adapters. Moreover, they continue to use the original pre-trained model to initialize the downstream model, leading to negligible changes in the model's generalization compared to the original model. In addition, there is still a lack of research investigating the consequences of integrating a multi-modal model into the updating procedure for both uni-modal and multi-modal tasks and the subsequent impacts it has on downstream tasks. In this paper, we propose an adapter-based two-stage learning paradigm, a multi-modal continual learning scheme that consists of experience-based learning and novel knowledge expansion, which helps the model fully use experience knowledge and compensate for novel knowledge. Extensive experiments demonstrate that our method is proficient for continual learning. It expands the distribution of representation upstream while also minimizing the negative impact of forgetting previous tasks. Additionally, it enhances the generalization capability for downstream tasks. Furthermore, we incorporate both multi-modal and uni-modal tasks into upstream continual learning. We observe that learning from upstream tasks can help with downstream tasks. \textit{Our code will be available at:} \url{https://github.com/lihong2303/ATLAS}.
\end{abstract}

\section{Introduction}
\label{sec:intro}

Continual learning provides a viable approach to enhance a model's capability by sequentially updating it with incoming tasks. In recent years, there has been an increased focus on multi-modal continual learning~\cite{gai2021multi,sun2020multimodal}, primarily driven by the rise of multi-modal foundation models. This approach tackles scenarios where incoming tasks can originate from diverse modalities, thereby receiving more and more attention in the research community. The intuition of involving multiple modalities is based on the human learning process, where successfully acquiring knowledge from one modality task significantly enhances proficiency in performing other tasks.  However, there is usually a trade-off between acquiring new knowledge and maintaining old knowledge in continual learning, where there will be a (catastrophic) forgetting of older information when learning new tasks. Moreover, continual learning in a multi-modal model is more challenging because it additionally requires the ability to update the model for each individual uni-modal task and may suffer from a mismatch between different modalities~\cite{liang2022mind}.

Recently, parameter-efficient finetuning has shown promise in alleviating catastrophic forgetting by learning a unique module for each task. This suits continual learning in multi-modal models, as multi-modal and uni-modal tasks often have different distributions, and generalizing to one can hurt the performance on others. Recent prompt-based studies~\cite{wang2022dualprompt,wang2022learning,smith2023coda,wang2022s} address continual learning in class-incremental and domain-incremental settings by learning prompts for each task. Similarly, the adapter-based study~\cite{srinivasan2022climb} learns individual adapter modules for each task. Prompt-based methods and adapter-based methods usually insert small, task-specific modules into the pre-trained models. However, neither method substantially expands the model's existing knowledge when encountering many tasks. Moreover, most research in multi-modal learning tends to explore the integration of different modalities encoders (e.g., text, image) to solve tasks that require understanding and processing multiple modal inputs. 
Therefore, the impact of employing a multimodal model~\cite{kim2021vilt} that jointly handles multimodal inputs for individual unimodal tasks remains relatively less explored in continual learning.

\begin{figure}[t]
  \centering
   \includegraphics[width=0.9\textwidth]{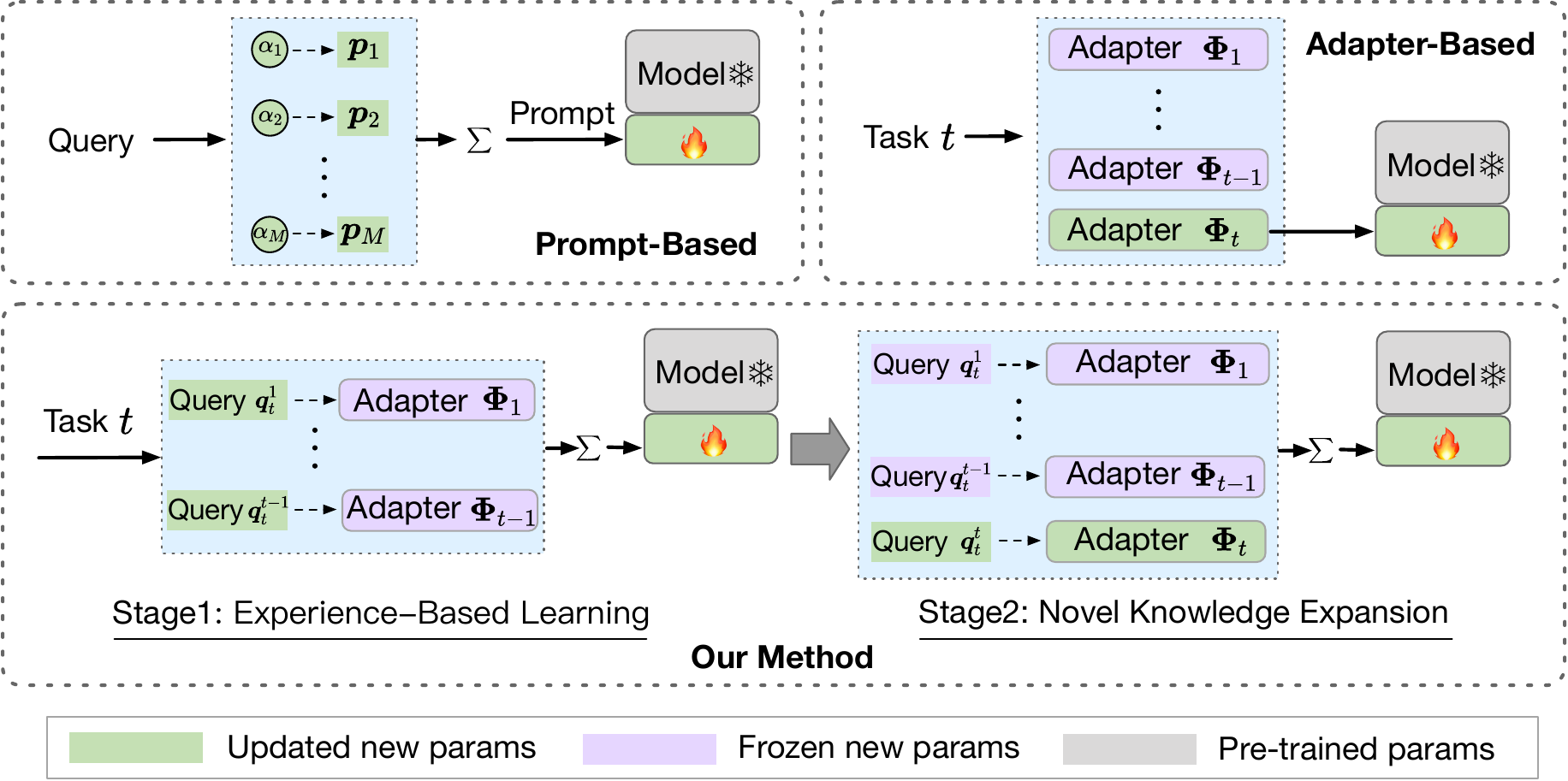}
   \caption{Prior prompt-based work involves a pool of key-value pairs used to select learnable prompts inserted in the model for learning instruction. Adapter-based work entails the individual learning of adapter modules for each task. Our methods propose a two-stage learning paradigm, which is a knowledge-incremental approach to constantly expanding knowledge when learning sequential tasks. {We train stage2 in the subsequent epochs after stage1 is completed.} }
   \label{fig:fig-teaser}
\end{figure}

Therefore, in this paper, we introduce \textbf{A}dapter-based Multi-modal Con\textbf{T}inual \textbf{L}earning with \textbf{A} Two-stage Learning \textbf{S}trategy (ATLAS). ATLAS follows a two-stage learning paradigm: experience-based learning and novel knowledge expansion (Figure~\ref{fig:fig-teaser}), utilizing multimodal models that directly process multimodal inputs.
The goal of experience-based learning is to learn how to effectively utilize previously-seen task knowledge when encountering new tasks. Based on experience-based learning, the novel knowledge expansion further compensates for the knowledge outside of the previously seen task.
This not only ensures that the knowledge is thoroughly explored for each task but also avoids the redundancy of knowledge between tasks sequentially.
Our method enhances the diversity of distribution and improves downstream generalization capability during upstream sequential tasks.
Moreover, we consider tasks that include both multi-modal and individual uni-modal tasks for both upstream and downstream processes.
This helps us analyze the influence of utilizing a multimodal model that jointly processes multimodal inputs for separate unimodal tasks.

The main contributions of this work can be summarized as follows:
(1) We introduce a two-stage learning paradigm for multi-modal continual learning to enhance the richness of distribution and improve the generalization capability;
(2) We propose a knowledge vector learning method that can combine the knowledge from different adapters based on the cosine similarity between the previous-seen tasks and the new task;
(3) We systematically analyze the effect of learning the multi-modal model on individual uni-modal tasks.

\section{Related work}
\label{sec:related-work}

\subsection{Continual Learning}

The methods of continual learning can be mainly classified into three categories: regularization-based \cite{kirkpatrick2017overcoming,aljundi2018memory,zenke2017continual,li2017learning,wang2021afec,dhar2019learning}, rehearsal-based \cite{wu2019large,prabhu2020gdumb,buzzega2020dark,wang2021memory,wang2021ordisco,wang2021triple}, and architecture-based \cite{serra2018overcoming,rusu2016progressive,wang2022coscl,yang2022continual}.
The regularization-based methods mitigate catastrophic forgetting \cite{mcclelland1995there} by stabilizing the modification of previous model parameters when learning new tasks.
The regularization in weight space penalizes changes in model parameters, while the prediction space penalizes changes in model predictions.
The rehearsal-based methods require a memory buffer to store previous task samples, which are utilized to recover previously encountered distributions and prevent catastrophic forgetting.
The architecture-based methods add task-specific architectures for each new task and expand the overall capacity of the model, which is significantly important for improving the capability to learn new knowledge.

In this paper, we focus on enhancing the richness of model representation and improving its generalization capability using architecture-based methods.

\subsection{Parameter-Efficient Finetuning for Continual Learning}
Recently, using pre-trained foundation models has increasingly become the workhorse for continual learning. Recent studies have introduced continual learning algorithms that are based on parameter-efficient networks, including prompts and adapters.
CLiMB~\cite{srinivasan2022climb} proposes that the adapter is a promising solution to avoiding catastrophic forgetting for task-incremental learning.
DualPrompt~\cite{wang2022dualprompt} and L2P~\cite{wang2022learning} construct a set of prompts and select the correlated prompt with input data, which is then inserted into the model to learn model instruction for class-incremental learning.
CODA-Prompt~\cite{smith2023coda} builds on the intuition above, which enhances the learning capacity of prompting by decomposing a learnable prompt into a set of prompt components.
S-Prompts~\cite{wang2022s} also learn prompts for new tasks but are designed for domain-incremental learning.
However, the adapter-based method learns a task-specific adapter, which lacks the capability to enhance the knowledge of the pre-trained model.
While the prompt-based methods learn the instructions for previous tasks, they still lack the capability to expand the knowledge of the pre-trained model.

\section{Preliminaries}

\subsection{Continual Learning}
Assume that we are given a multi-modal model $\mathcal{M}_0$, which is pre-trained on an initial task. The primary focus is on task-incremental continual learning, which involves the sequential arrival of tasks. In this setting, the model only has access to the current training data and is not able to access any data from previous tasks. At time $t$, the model updates its parameters based on the loss function $\mathcal{L}_t$ and training data $\mathcal{D}_t$. We assume that a multi-modal model undergoes continuous training on a sequence of $T$ distinct \emph{upstream tasks}.

An effective continual learning paradigm should exhibit a minimal amount of forgetting about previously encountered tasks. This means that after training on subsequent tasks, the average accuracy of a previously trained task should remain close to its initial accuracy. In other words, the model should retain its knowledge and performance on earlier tasks even as it learns new ones. Another crucial criterion for a reliable continual learning method is that its generalization ability on \emph{downstream tasks} (which have no intersections with upstream tasks) should improve as the training process progresses. In other words, the method should exhibit an increasing capability to generalize and perform well on unseen tasks as it continues to learn and accumulate knowledge from (upstream task) learning. This indicates that the model becomes more proficient in applying its learned knowledge to a diverse range of unseen tasks over time.

Regardless of whether we consider tasks in the upstream or downstream context, our investigation encompasses a diverse set of tasks that involve both multiple modalities and uni-modalities. By doing so, we aim to understand the effects of training a multi-modal model on tasks that rely on individual modalities, while also exploring the potential benefits that arise for both downstream multi-modal tasks and tasks focused on individual modalities.

\subsection{Adapters}

The adapter, introduced in the paper by \cite{houlsby2019parameter}, presents an alternative approach for adapting a pre-trained model to a specific task. In contrast to resource-intensive fine-tuning, an adapter module consists of a considerably smaller number of parameters compared to the original model. Despite this reduction, adapter-based models generally achieve performance that is comparable to fine-tuning the entire model.

In practical, the parameters of the pre-trained model stay fixed, while the adapter modules are trained for a given task. This enables quick adaptation to new tasks without the need for extensive retraining of the entire model.

In this work, the initialized model for upstream continual learning is a pre-trained vision-language model $\mathcal{M}_{0}$. 
For each task at time $t$ and its corresponding dataset $\mathcal{D}_{t}$, the model $\mathcal{M}_{t}$ is initialized from the previous upstream updated step checkpoint $\mathcal{M}_{t-1}$.
We continuously learn a task-specific adapter module $\boldsymbol{\Phi}_{t}$ for each new task based on all previous task modules as:
\begin{equation}
    {\boldsymbol{\Phi}_{t}} = \argmin_{{\boldsymbol{\Phi}'_{t}}}  \mathcal{L}_t\left({\boldsymbol{\Phi}'_{t}} \mid \mathcal{D}_{t} ; \mathcal{M}_0, {\boldsymbol{\Phi}_{0}},\ldots,{\boldsymbol{\Phi}_{t-1}}\right),
\end{equation}
where $\mathcal{L}_t$ refers to the loss function used for the $t$-th task and ${\boldsymbol{\Phi}_{0}} = \mathbf{0}$.

\begin{figure}[t]
  \centering
   \includegraphics[width=0.87\linewidth]{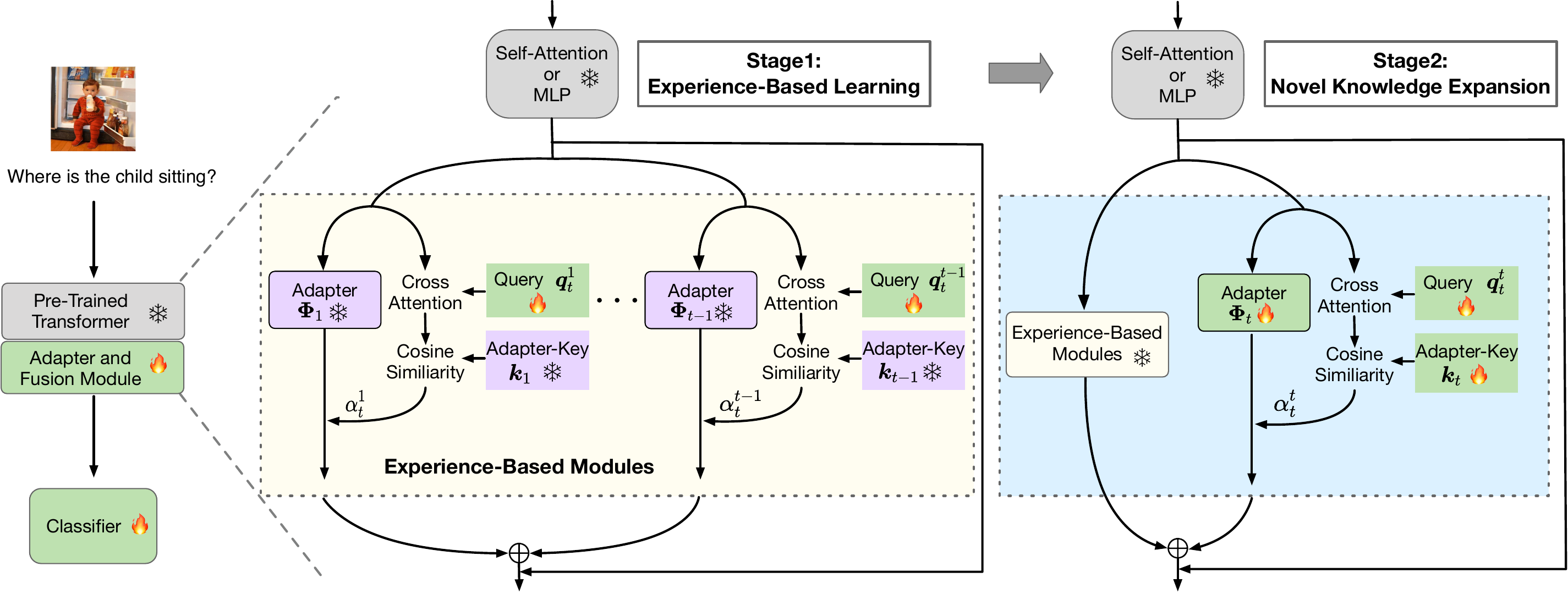}
   \caption{Schematic Figure of the two-stage learning paradigm. For the $t$-th task, experience-based learning optimizes the parameters of Query to control the degree of involvement of previously saved knowledge in the new task. Based on Novel Knowledge Expansion, we compensate for the knowledge not exist in previously saved knowledge to ensure the full exploration of new task knowledge. The meaning of the modules' color is similar to Figure \ref{fig:fig-teaser}.
   }
   \label{fig:figure-methods}
\end{figure}

\section{Method}
\label{sec:method}

While using adapters has shown a promising way to alleviate catastrophic forgetting, the existing works lack the investigation of enhancing the generalization ability of the foundation model for downstream tasks.
Specifically, the adapter-based method~\cite{srinivasan2022climb} usually trains an individual adapter module for each task during upstream training. 
However, the original model is still used for initialization when seeing new tasks, and the upstream learned knowledge is not involved in the new incoming task.
One potential way is to leverage the existing knowledge in the upstream learned task adapters and the original pre-trained model for the incoming task.
Intuitively, to tackle a novel task, the model needs to leverage prior knowledge from the initialized checkpoint and assimilate new knowledge specific to the current task.
Additionally, we desire an approach that constantly improves the modal capability, which we conjecture is helpful for the generalization capability for adapting to new tasks.

We propose a two-stage learning paradigm for multi-modal continual learning that utilizes adapters to expand knowledge, enabling the model to continuously enhance its generalization ability for new tasks.
Rather than training a separate adapter for each task, we introduce an adapter weighting method that incorporates the knowledge of different adapters to tackle new tasks.
This allows us to enhance the capability of the foundation model and capture different task-specific knowledge, which can be beneficial for new incoming tasks.

In the following, we introduce a two-stage learning paradigm in Section~\ref{two-stage learning paradigm}. Specifically, we discuss how to calculate the knowledge coefficient vector in Section \ref{knowledge vector learning}. We describe each stage of our method in detail in Sections~\ref{experience-based learning} and~\ref{novel knowledge expansion}. Finally, we introduce the additional strategy and optimization function in Section~\ref{strategy and optimization}. We abbreviate the learnable classifier's parameter when writing optimization problems for notational simplicity.

\subsection{Two-Stage Learning Paradigm}
\label{two-stage learning paradigm}

Crucial to expanding knowledge is utilizing the knowledge involved in the previous task adapters. Empirical studies~\cite{von2021informed} propose that various knowledge sources, representations, and integration methods can boost the learning of new knowledge.
This is intuitive for advanced intelligent beings like humans, who naturally associate previously learned knowledge when learning new knowledge.
However, in the current process of continual learning for the multi-modal model, the new downstream task still relies on the original pre-trained model for initialization, and the knowledge from the previous task is not incorporated into the learning process for the new task.
They independently learn a new architecture for the incoming task, resulting in redundant knowledge among task-specific architectures and limiting the reserve of knowledge for new tasks.
This leads to an urgent need for methods that effectively utilize experience-based learning for incoming tasks.

To better illustrate the working process of the paradigm, we will first briefly present the \emph{schematic} two-stage learning paradigm consisting of experience-based learning and novel knowledge expansion as follows, detailed implementations will be discussed in section \ref{knowledge vector learning}, \ref{experience-based learning}, \ref{novel knowledge expansion}, and \ref{strategy and optimization}. Figure~\ref{fig:figure-methods} presents the illustration of the proposed two-stage learning paradigm.
Experience-based learning hopes to fully use the previous comprehensive knowledge to tackle new tasks. 
Such knowledge may include original pre-trained model knowledge $\mathcal{M}_{0}$ as well as task-specific knowledge $\boldsymbol{\Phi}_{i}$ gained from previously observed upstream tasks. Moreover, we introduce the knowledge coefficient vector to control the degree of participation of different knowledge in dealing with new tasks.
For the upstream $t$-th task ($t>1$), we optimize the knowledge coefficient within the framework of experience-based learning as follows:
\begin{equation}
    {\boldsymbol{\alpha}_{-t}} = \mathop{\arg\min} \limits_{{ \bar{\boldsymbol{\alpha}}_{-t}}}  \mathcal{L}_t\left({\bar{\boldsymbol{\alpha}}_{-t}} \mid \mathcal{D}_{t} ; \mathcal{M}_0, {\boldsymbol{\Phi}_{1}},\ldots,{\boldsymbol{\Phi}_{t-1}}\right),
\end{equation}
where $\bar{\boldsymbol{\alpha}}_{-t} = (\bar{\alpha}^1_t, \cdots, \bar{\alpha}^{t-1}_{t})$.
Then, we decompose the new task using previously acquired knowledge. As there may be some additional knowledge beyond what was previously acquired.
To further fully explore the knowledge involved in a new task, we introduce an additional knowledge expansion stage to learn compensation knowledge.
For the upstream $t$-th task, we learn a new task-specific module $\boldsymbol{\Phi}_{t}$ and corresponding knowledge coefficient $\alpha^{t}_{t}$ within the novel knowledge expansion stage to expand the additional knowledge as:
\begin{equation}
    \boldsymbol{\Phi}_{t}, \alpha^{t}_{t} = \argmin_{\boldsymbol{\Phi}'_{t}, \bar{\alpha}^{t}_{t}} \mathcal{L}_t \left( \boldsymbol{\Phi}'_{t}, \bar{\alpha}^{t}_{t} \mid \mathcal{D}_t; \mathcal{M}_0, \boldsymbol{\Phi}_1, \ldots, \boldsymbol{\Phi}_{t-1}, \alpha_{-t} \right)
\end{equation}
We obtain the knowledge coefficient as $\boldsymbol{\alpha}_t = (\boldsymbol{\alpha}_{-t}, \alpha^t_t)$.

\subsubsection{Knowledge Vector Learning}
\label{knowledge vector learning}

Based on a set of learned task-specific adapters, we desire to produce the knowledge coefficient vector $\boldsymbol{\alpha}$ via correlating the sample-level latent representation with the corresponding key of the adapter.
The underlying intuition is that the involvement degree of each adapter to a new task is determined by the similarity between the sample-level latent representation $\boldsymbol{h}(\boldsymbol{x})$ and corresponding indicator $\boldsymbol{k}_i$ for the $i$-th adapter.
More importantly, the different adapter keys need to connect with the different aspects of latent representation, which indicates the demand for each adapter to be involved in the current task.
However, the network's latent representation for similarity computation requires a key with the same dimension, which is more computationally expensive.
Additionally, it is challenging to focus on the knowledge corresponding to the adapter.

To solve the above challenges, we propose to compute a sample-level description vector $\boldsymbol{d}_{i}(\boldsymbol{x_{t}})$ for $t$-th task input $\boldsymbol{x_{t}}$ with respect to the $i$-th adapter, which captures the different components of the latent representation.
It boosts our method's computation efficiency and enhances the accuracy of making use of different adapter modules. 
Then, for each adapter $\boldsymbol{\Phi}_{i}$, we assign a learnable reference vector $\boldsymbol{k}_{i} \in \mathbb{R}^D$ as its abstract description.

Drawing inspiration from the query-based feature extraction method~\cite{alayrac2022flamingo}, we introduce a learnable query to capture the sample-level description corresponding to each adapter in the latent representation of the network. For any sample $\boldsymbol{x}_{t} \in \mathcal{D}_t$, we compute cross-attention between the latent representation $h(\boldsymbol{x}_{t})$ within the $t$-th task input $\boldsymbol{x}_{t}$ and a learnable query $\boldsymbol{q}^{i}_{t} \in \mathbb{R}^D$ that corresponds to each adapter to produce the sample-level description as follows:
\begin{equation}
    \boldsymbol{d}_i(\boldsymbol{x}_{t}) = \text{CrossAtt}(\boldsymbol{h}(\boldsymbol{x}_{t}), \boldsymbol{q}^{i}_t),
\end{equation}
where CrossAtt refers to the computation of cross-attention. $\boldsymbol{d}_i(\boldsymbol{x}_{t}) \in \mathbb{R}^D$ is the sample-level description for $i$-th task input corresponding to adapter $\boldsymbol{\Phi}_{t}$.
Notice that our cross-attention does not involve the projection operation and only utilizes the multiplying operation, which helps the multi-modal model to alleviate catastrophic forgetting.

Then we calculate the cosine similarity between the description vector $\boldsymbol{d}_i(\boldsymbol{x_{t}})$ and the reference vector $\boldsymbol{k}_i$ as:
\begin{equation}
    \alpha^i_{t} = \langle \boldsymbol{d}_i(\boldsymbol{x}_{t}), \boldsymbol{k}_i\rangle,
\end{equation}
where $\alpha^i_{t}$ is the weighting coefficient for the $i$-th adapter in the $t$-th task.
This captures the overlapping of the task input $\boldsymbol{x_{t}}$ and the adapter $\boldsymbol{\Phi}_{i}$. 
We use $\alpha^i_{t}$ as the weight of the output of $\boldsymbol{\Phi}_i$ for the $t$-th task adapter weighting. The loss on the whole dataset $\mathcal{D}_t$ is just the average loss of all the samples.

\subsubsection{Stage1: Experience-Based Learning}
\label{experience-based learning}

Experience-based learning solves new tasks by decomposing task knowledge into previously saved knowledge.
In our work, we adopt the adapter module $\boldsymbol{\Phi}_{i}$ as the basic structure for knowledge \emph{increment} in our two-stage learning paradigm.
Specifically, for the upstream $t$-th task, we attain adapter-based knowledge through insert a weighted summation and form a decomposed adapter into the training to effectively utilize the knowledge of the previous task adapter: $\hat{\boldsymbol{\Phi}}^t_{\text{Stage 1}} = \sum^{t-1}_{i=1}\alpha^{i}_{t} \hat{\boldsymbol{\Phi}}_{i}$,
where $\hat{\boldsymbol{\Phi}}_{i}$ is the output of the $i$-th adapter module. The $\alpha^{i}_{t}$ is the weighting coefficient for the $i$-th adapter that determines the degree of each adapter involved in the current tasks, {see Figure~\ref{fig:figure-methods} for better understanding}.

Thus, the optimization problem for experience-based learning can be described as follows:
\begin{equation}
  \boldsymbol{q}_{-t}, \boldsymbol{K}_{-t} = \argmin_{\boldsymbol{q}'_{-t}, \boldsymbol{K}'_{-t}} \mathcal{L}_t \left( \boldsymbol{q}'_{-t}, \boldsymbol{K}'_{-t} \mid \mathcal{D}_t; \mathcal{M}_0, \boldsymbol{\Phi}_1, \ldots, \boldsymbol{\Phi}_{t-1} \right)
\end{equation}
where $\boldsymbol{q}_{-t} = [\boldsymbol{q}^{1}_{t}, \ldots, \boldsymbol{q}^{t-1}_{t}]$ is the learned collection of task-specific queries for previously seen $t-1$ tasks and $\boldsymbol{K}_{-t} = [\boldsymbol{k}_{1}, \ldots, \boldsymbol{k}_{t-1}]$ is the learned set of adapter description vectors.

\subsubsection{Stage2: Novel Knowledge Expansion}
\label{novel knowledge expansion}

With experience-based learning as a foundation, we are ready to expand new knowledge that has not been covered in the previous adapters.
This ensures that we can fully learn the knowledge carried by new tasks and continuously expand the non-redundant knowledge of the foundation model.
The model is \emph{initialized} using experience-based generalization learning. This includes the original pre-training model $\mathcal{M}_0$, a collection of previous task adapters $\boldsymbol{\Phi}_{1}, \ldots, \boldsymbol{\Phi}_{t-1}$ and their corresponding reference vector $\boldsymbol{k}_{1}, \ldots, \boldsymbol{k}_{t-1}$, as well as queries $\boldsymbol{q}^{1}_{t}, \ldots, \boldsymbol{q}^{t-1}_{t}$ specific to $t$-th upstream task.
Then we introduce a set of learnable parameters: adapter $\boldsymbol{\Phi}_{{t}}$ and its corresponding adapter reference vector $\boldsymbol{k}_{t}$, along with a query $\boldsymbol{q}^{t}_{t}$ to compensate for new knowledge.

Similar to the experienced-based learning stage, we use $\hat{\boldsymbol{\Phi}}^t_{\text{Stage 2}} = \sum^{t}_{i=1}\alpha^{i}_{t} \hat{\boldsymbol{\Phi}}_{i}$ in the training of the novel knowledge expansion, then the novel knowledge expansion is an optimization problem as follows:
\begin{equation*}
    {\boldsymbol{\Phi}_{t}},{\boldsymbol{q}^{t}_{t}},{\boldsymbol{k}_{t}} =  \mathop{\arg\min} \limits_{{\boldsymbol{\Phi}'_{t}},{\bar{\mathbf{q}}^{t}_{t}},{\boldsymbol{k}'_{t}}}  \mathcal{L}_t 
 ({\boldsymbol{\Phi}'_{t}},{\bar{\boldsymbol{q}}^{t}_{t}},\boldsymbol{k}'_{t} \mid \mathcal{D}_{t} ; \mathcal{M}_0,  \nonumber
      {\boldsymbol{\Phi}_{1}}, \ldots, {\boldsymbol{\Phi}_{t-1}}, {\boldsymbol{q}_{-t}},{\boldsymbol{K}_{-t}} ).
\end{equation*}

The performance of the above novel knowledge expansion reflects how much the multi-modal model attains knowledge for the current task.
Additionally, the compensated new task knowledge may be useful for future tasks.

\subsubsection{Orthogonality \& Optimization}
\label{strategy and optimization}

The two-stage learning paradigm is focused on improving the richness of distribution and enhancing generalization capability.
For the $t$-th task, its task-specific query $\boldsymbol{q}_{t}=[\boldsymbol{q}^1_{t}, \cdots, \boldsymbol{q}^t_{t}]$ and adapter-specific key $\boldsymbol{K}_{t} = [\boldsymbol{k}_{1}, \ldots, \boldsymbol{k}_{t}]$ are used to determine the weighting value.
The empirical study~\cite{smith2023coda} proposes that orthogonal vectors have less interference between existing and new knowledge.
In our work, we add the orthogonal constraint to $\boldsymbol{q}_{t}$ and $\boldsymbol{K}_{t}$.
Specifically, we introduce the orthonormal regularization as:
\begin{equation*}
    \mathcal{L}_{ortho} (\boldsymbol{X}) = \|\boldsymbol{X}^\top \boldsymbol{X} - \boldsymbol{I}\|_F,
\end{equation*}
where $\|\cdot\|_F$ represents Frobenius norm and $\boldsymbol{X}$ is $\boldsymbol{q}_{t}$ or $\boldsymbol{K}_{t}$.

Combined with task-specific loss $\mathcal{L}_{t}$, the total loss is defined as:
\begin{equation}
    \quad \mathcal{L}_{t} + \lambda\left(\mathcal{L}_{ortho}\left(\boldsymbol{q}_{t}\right) + \mathcal{L}_{ortho}\left(\boldsymbol{K}_t\right)\right),
\end{equation}
where $\lambda$ is a hyper-parameter to control the degree of orthogonal constraint.

\section{Experiments}
\label{sec:exp}

\begin{figure*}[t]
  \centering
   \includegraphics[width=1.0\textwidth]{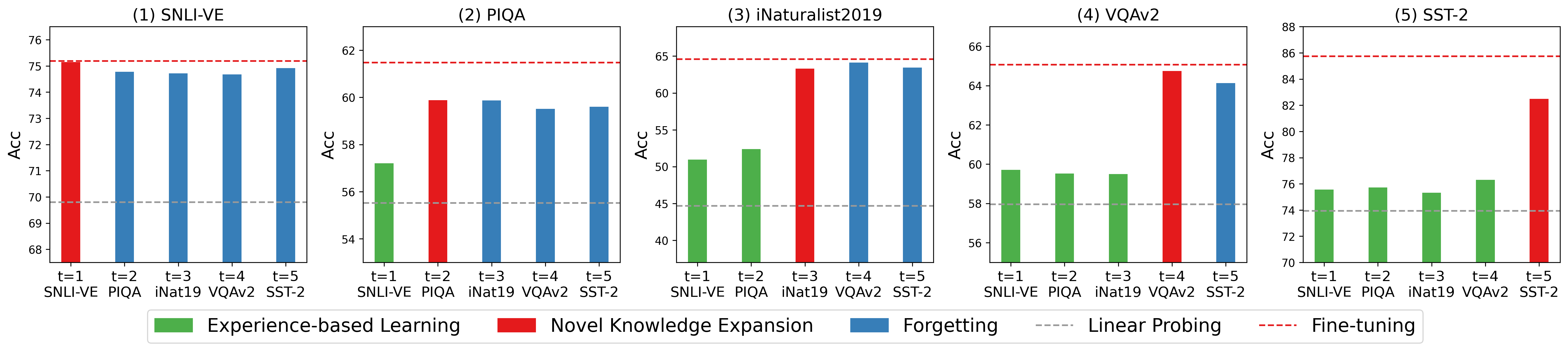}
   \caption{Learning process of upstream continuous learning. {Linear Probing} is the lower bound of continual learning while {Fine-tuning} serves as the upper bound. We estimate experience-based learning before novel knowledge expansion in each task, and estimate forgetting afterward.} 
   \label{fig:result-figure1}
\end{figure*}

\subsection{Experimental Setting}
 
In this paper, we apply our adapter-based two-stage learning paradigm approach to task-incremental multi-modal model continual learning that covers upstream continual learning and downstream generalization, both of which contain multi-modal and uni-modal tasks.
For upstream continual learning, our method is compared with the adapter method proposed in CLiMB~\cite{srinivasan2022climb}, which learns a task-specific adapter for each task.
To measure downstream generalization, we estimate the experience-based learning capability and low-shot learning ability for the upstream updated model.

For upstream continual learning, compared with CLiMB~\cite{srinivasan2022climb} which only uses four multi-modal upstream tasks, we carry out experiments on five popular datasets to enhance the experiment setting, including the multi-modal SNLI-VE~\cite{xie2019visual}, multi-modal VQAv2~\cite{goyal2017making}, language PIQA~\cite{bisk2020piqa},
language-only SST-2~\cite{socher2013recursive}, 
and image-only iNaturalist2019~\cite{van2018inaturalist}.
For downstream generalization, we access four different datasets, including multi-modal VCR~\cite{zellers2019recognition}, multi-modal NLVR2~\cite{suhr2018corpus}, language CommonsenseQA~\cite{talmor2018commonsenseqa}, and image Places365~\cite{lopez2020semantic}.
The detailed information regarding the details of the dataset and experimental setup refers to the supplementary material.

For the low-shot experiments, we adopt the same samples as CLiMB~\cite{srinivasan2022climb}. We randomly selected 2048, 32, and 1024 examples per class for NLVR2~\cite{suhr2018corpus}, Places365~\cite{lopez2020semantic}, and CommonsenseQA~\cite{talmor2018commonsenseqa}, respectively.
For the multiple-choice VCR~\cite{zellers2019recognition} tasks, we use 5\% training data in the experiment.
Following CLiMB~\cite{srinivasan2022climb}, we use the pre-training ViLT~\cite{kim2021vilt} to initialize the upstream learning model.

\subsection{Upstream Continual Learning}

In this subsection, we focus on the training process of upstream continual learning and demonstrate that our two-stage learning paradigm is an effective continual learner without forgetting while having powerful upstream experience-based learning capability.
In upstream continual learning, we use a fixed upstream task order: Multi-First (SNLI-VE$\rightarrow$PIQA$\rightarrow$iNaturalist2019$\rightarrow$VQAv2$\rightarrow$SST-2).

In our experiments, we use the trained parameters of the task-specific query for additional adapter modules, this allows us to analyze the impact of the current task update on subsequent tasks. Notably, CLiMB~\cite{srinivasan2022climb} can be seen as a special case of our method by setting the task-specific query of additional adapter to zero vector, which produces that the corresponding adapter coefficient is zero. The {Forgetting} displayed in our results is similar to prior work like CLiMB~\cite{srinivasan2022climb}. See Supplementary materials for more detailed discussions.

Figure~\ref{fig:result-figure1} presents the results of the training process during upstream continual learning.
Following the fixed task order, we estimate the experience-based learning performance on unseen tasks while estimating forgetting on tasks that have been previously learned.
Experience-based learning optimizes additional task-specific parameters compared to linear probing updates, which is \emph{negligible} given that each task-specific parameter for one adapter is 18.4k compared to 3.1M adapter parameters.
Novel knowledge expansion learns task-specific adapters based on experience-based learning in the current task.
We only learn the task-specific query for the newly added adapters when estimating forgetting.
To better estimate the performance of our algorithm, we introduce the linear probing as our lower bound and the fine-tuning update as our upper bound.

Our approach is an effective continuous learner, achieving performance close to fine-tuning across the entire dataset and demonstrating a strong ability to mitigate catastrophic forgetting.
Adding the compensation knowledge of tasks after the current task does not improve performance, except for a subtle improvement in the iNaturalist2019 dataset.
It is intuitive that our two-stage learning paradigm fully explores the knowledge in each task.
Experience-based learning decomposes the new task into knowledge from previous tasks, while novel knowledge expansion compensates for the knowledge beyond previously saved knowledge.
The improvement of $\textit{experience-based learning}$ over $\textit{linear probing}$ reflects the additional task knowledge that assists in new task learning.

\subsection{Downstream Generalization}

The capability of downstream generalization reflects the richness of distribution and the adaptability of a new task.
In the following, we access the downstream learning capability based on experience-based learning and the low-shot capability in each step of upstream continual learning.

\begin{table}[t]
    \centering
    \setlength{\tabcolsep}{1.0mm}
    \resizebox{0.65\linewidth}{!}{%
    \begin{tabular}{clcccc>{\cellcolor{cellcolor}}c}
    \toprule
    \multicolumn{2}{c}{Downstream Task} & {VCR} & {NLVR2} & {CommQA} & {Places365} & {Avg.} \\
    \midrule
    \multicolumn{2}{c}{Linear Probing} & 46.74 & 63.53 & 28.75 & 44.23 & 45.81 \\
    \midrule
    \multirow{5}{*}{
        \begin{tabular}{@{}c@{}}
            \rotatebox[origin=c]{90}{{Adapter~\cite{srinivasan2022climb}}}
        \end{tabular}
    } &
    {SNLI-VE} & 47.33 & 63.92 & 28.67  & 43.80 & 45.93 \\
    & {PIQA} & 45.81 & 62.69 & 29.22 & 43.94 & 45.42 \\
    & {iNat2019} & 47.16 & 61.94 & 25.92 & 46.95 & 45.49 \\
    & {VQAv2} & 46.86 & 63.47 & 28.58 & 45.81 & 46.18 \\
    & {SST-2} & 47.23 & 63.35 & 28.79 & 42.88 & 45.56 \\
    \midrule
    \multirow{5}{*}{
        \begin{tabular}{@{}c@{}}
            \rotatebox[origin=c]{90}{{Ours}}
        \end{tabular}
    } &
    {SNLI-VE {\footnotesize(t=1)}} & 48.87 & 65.10 & 29.45 & 47.36 & 47.70 \\
    & {PIQA {\footnotesize(t=2)}} & 48.82 & 65.39 & 30.52 & 48.19 & 48.23 \\
    & {iNat2019 {\footnotesize(t=3)}} & 48.97 & 65.12 & 30.10 & 49.12 & 48.33 \\
    & {VQAv2 {\footnotesize(t=4)}} & 49.40 & 65.47 & 31.69 & 49.58 & 49.04 \\ 
    & {SST-2 {\footnotesize(t=5)}} & 49.62 & 65.56 & 32.06 & 49.27 & 49.09 \\
    \bottomrule
    \end{tabular}}
    \caption{Linear Probing accuracy of the upstream updated model at each step in downstream tasks. {Linear probing} is the accuracy of generalizing the original pre-trained model. {Adapter} is the generalization accuracy of the CLiMB~\cite{srinivasan2022climb} method. $\textit{Ours}$ is the downstream experience-based learning accuracy of our method.}
    \label{tab:downstream_experience_learning}
\end{table}
Table~\ref{tab:downstream_experience_learning} shows that our method significantly improves the capability of downstream experience-based learning compared to prior works.
It also maintains a constantly increasing trend as knowledge constantly expands.
For prior work CLiMB~\cite{srinivasan2022climb}, the experience-based learning capability of the model requests the use of the full individual adapter.
If there is no relevance between the upstream task and the downstream task, the generalization ability of the downstream model will greatly decrease.
For instance, the upstream iNaturalist image dataset adapter-based model adapts to the downstream natural language CommonsenseQA dataset, with accuracy decreasing from 28.75 to 25.92.
Opposed to CLiMB, our method can learn the effective combination of different adapter knowledge.
If a downstream task is unrelated to one of the upstream tasks, our method can learn a knowledge coefficient close to zero for this adapter.
For example, the downstream VCR dataset is unrelated to the upstream PIQA dataset, so the accuracy remains unchanged when adding PIQA adapter knowledge.

\begin{table}[t]
    \centering    
    \begin{minipage}{0.45\linewidth}
        \centering
    \setlength{\tabcolsep}{1.0mm}
    \resizebox{\linewidth}{!}{%
    \begin{tabular}{lccccc>{\cellcolor{cellcolor}}c}
    \toprule
    \multicolumn{2}{c}{Downstream Task}   & {VCR} & {NLVR2} & {CommQA} & {Places365} & {Avg.} \\
    \midrule
    \multicolumn{2}{c}{Linear Probing~\cite{srinivasan2022climb}} & 40.14 & 60.45 & 27.89 & 27.97 & 39.09 \\
    \midrule
    \multirow{5}{*}{
        \begin{tabular}{@{}c@{}}
            \rotatebox[origin=c]{90}{{Ours}}
        \end{tabular}
    } &{SNLI-VE} {\footnotesize(t=1)}  & 41.49 & 61.07 & 28.31 & 32.75 & 40.91 \\
    &{PIQA} {\footnotesize(t=2)} & 41.89 & 61.68 & 29.70 & 32.98 & 41.56 \\
    &{iNat2019} {\footnotesize(t=3)} & 42.23 & 61.43 & 29.38 & 33.63 & 41.67 \\
    &{VQAv2} {\footnotesize(t=4)} & 42.06 & 62.12 & 30.02 & 34.51 & 42.18 \\
    &{SST-2} {\footnotesize(t=5)} & 42.32 & 62.66 & 30.87 & 34.42 & 42.57 \\
    \bottomrule
    \end{tabular}}
    \caption{Low-shot Linear Probing accuracy of upstream updated model at each step on downstream tasks. 
    }
    \label{tab:few-shot_learning}
    \end{minipage}
    \begin{minipage}{0.45\linewidth}
        \centering
    \setlength{\tabcolsep}{1.0mm}
    \resizebox{\linewidth}{!}{%
    \begin{tabular}{lccccc>{\cellcolor{cellcolor}}c}
    \toprule
    \multicolumn{2}{c}{Downstream Task}   & {VCR} & {NLVR2} & {CommQA} & {Places365} & {Avg.} \\
    \midrule
    \multicolumn{2}{c}{Adapter} & 42.69 & 59.74 & 27.94 & 35.17 & 41.39 \\
    \midrule
    \multirow{5}{*}{
        \begin{tabular}{@{}c@{}}
            \rotatebox[origin=c]{90}{{Ours}}
        \end{tabular}
    } &{SNLI-VE }{\footnotesize(t=1)}  & 43.11 & 62.61 & 28.98 & 35.60 & 42.58 \\
    &{PIQA }{\footnotesize(t=2)} & 44.06 & 62.32 & 32.37 & 35.22 & 43.49 \\
    &{iNat2019 }{\footnotesize(t=3)} & 44.14 & 62.13 & 32.29 & 35.84 & 43.60 \\
    &{{VQAv2 }}{\footnotesize(t=4)} & 43.99 & 62.59 & 32.60 & 36.05 & 43.81 \\
    &{{SST-2}} {\footnotesize(t=5)} & 44.35 & 63.12 & 33.07 & 35.86 & 44.10  \\
    \bottomrule
    \end{tabular}}
    \caption{Low-shot adapter learning accuracy of upstream updated model at each step on downstream tasks.    
    }
    \label{tab:few-shot_learning-adapter}
    \end{minipage}
\end{table}

\begin{figure*}[t]
  \centering
   \includegraphics[width=1.0\textwidth]{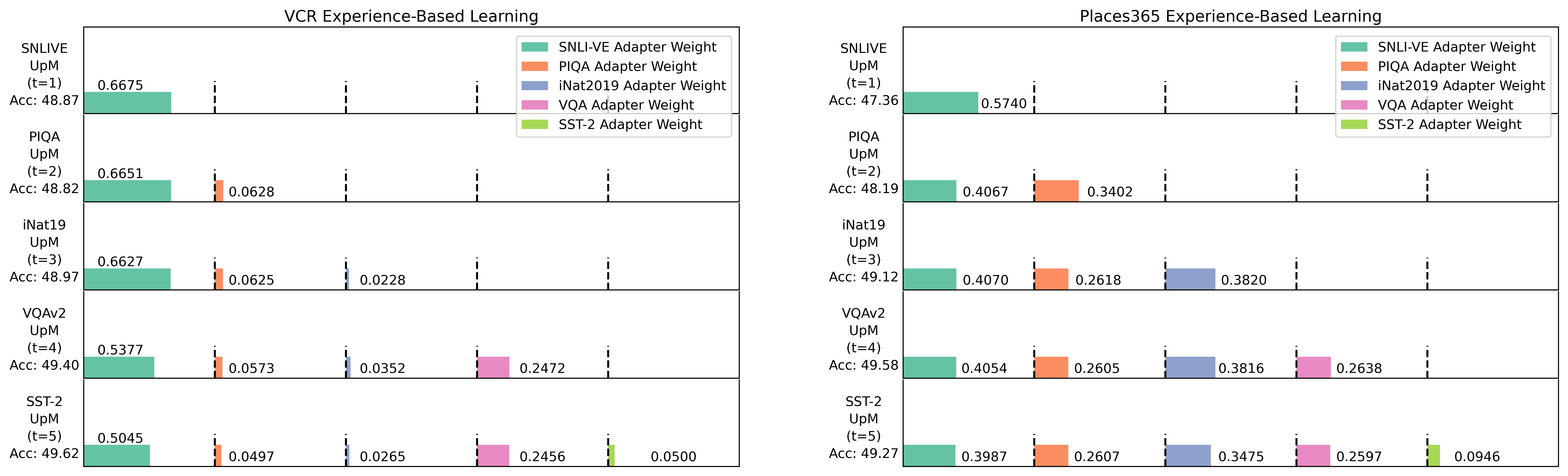}
   \caption{Accuracy and knowledge coefficient comparison of experience-based learning at each step of upstream continual learning on the VCR~\cite{zellers2019recognition} and Places365~\cite{lopez2020semantic} datasets. The length of each bar is a knowledge coefficient corresponding to each adapter.}
   \label{fig:experience-learning}
\end{figure*}

Table~\ref{tab:few-shot_learning} and~\ref{tab:few-shot_learning-adapter} present the result of downstream low-shot capability.
With low-shot samples, downstream experience-based learning also significantly enhances the model's performance, and as knowledge expands, the model's performance continues to increase consistently.
It is noticed that with low-shot samples, the adapter-learning performance is lower than linear probing in some cases, i.e., NLVR2.
One possible reason is that adapter learning overfits low-shot samples since adapter learning has more parameters than linear probing.
This reflects that low-shot samples contain less task knowledge.
Experience-based learning decomposes the new task knowledge over previously saved knowledge, improving the adaptation of the pre-training models to new tasks.
Combining our two-stage learning paradigm, downstream adapter-based novel knowledge expansion has enhanced performance compared to task-specific adapter learning. This indicates that our method is data-efficient with low-shot samples as our method ensures non-overlapping among different task knowledge, which can release more capability to learn new task knowledge.

We next show the impact of expanding multi-modal and uni-modal task knowledge on downstream tasks.
Figure~\ref{fig:experience-learning} compares the accuracy and knowledge coefficient at each step of upstream continual learning on the VCR and Places365 datasets, respectively.
We discovered that empirical-based learning effectively \emph{utilizes adapter knowledge according to the similarity between new tasks and adapters}.
In addition, we observe that the expansion of knowledge over multi-modal tasks typically enhances performance in both uni-modal and multi-modal tasks. However, the expansion of knowledge from uni-modal tasks offers little assistance for multi-modal tasks.
Intuitively, language task knowledge does not always translate to improvements in image tasks. Our experiments have shown that in certain cases, knowledge from uni-modal tasks can benefit uni-modal tasks involving different modalities. For example, knowledge of PIQA language tasks enhances performance on Places365 tasks. We suspect that knowledge from a single-modal adapter contributes to the utilization of multi-modal models for single-modal tasks.
Furthermore, we have observed that the utilization of specific task knowledge is progressively diminishing as the saved knowledge continues to grow. We attribute this phenomenon to the presence of low-level competition in the integration of diverse tasks.

\section{Conclusion}
We propose a two-stage learning strategy for multi-modal continual learning: experience-based learning and novel knowledge expansion. Experience-based learning leverages old knowledge to decompose the task and fully utilize it. Novel knowledge expansion compensates for knowledge that goes beyond the saved knowledge. Our method continuously enhances the distribution of the updated model and improves experience-based learning and low-shot capability in downstream tasks. Analyzing different modalities enhances understanding of multi-modal continual learning. Future work includes applying our method to backbones like CLIP~\cite{radford2021learning} and exploring class-incremental learning settings~\cite{wang2022dualprompt, wang2022learning, smith2023coda}.











\bibliography{arxiv}
\bibliographystyle{plainnat}

\clearpage
\appendix
\begin{center}
    \Large \textbf{Appendix}\\[0.5cm]
\end{center}

\section{Implementation Details}

\subsection{Two-Stage Learning Paradigm Algorithm}

Algorithm~\ref{alg:TLAS} shows the details of our Two-Stage Learning Paradigm, including Experience-Based Learning and Novel Knowledge Expansion. For each time step t, Experienced-Based Learning aims to discover new task knowledge in already completed tasks knowledge. While Novel Knowledge Expansion compensates for the extra knowledge of a new task beyond the old task's knowledge.

\begin{algorithm}
    \caption{Two-Stage Learning Paradigm}
    \begin{algorithmic}[1]\label{alg:TLAS}
        \STATE \textbf{Input:} Upstream task sequence $1,\ldots , T$, pre-trained vision-language model $\mathcal{M}_{0}$, time step $t$, training data $\mathcal{D}_{t}$, loss function $\mathcal{L}_t$, task-specific adapter module $\boldsymbol{\Phi}_{t}$, old $t-1$ tasks task-specific queries $\boldsymbol{q}_{-t} = [\boldsymbol{q}^{1}_{t}, \ldots, \boldsymbol{q}^{t-1}_{t}]$, old $t-1$ tasks adapter description vectors $\boldsymbol{K}_{-t} = [\boldsymbol{k}_{1}, \ldots, \boldsymbol{k}_{t-1}]$, task $t$ query $\boldsymbol{q}^{t}_{t}$, task $t$ adapter description vector $\boldsymbol{k}_{t}$, $t$ tasks task-specific queries $\boldsymbol{q}_{t} = [\boldsymbol{q}^{1}_{t}, \ldots, \boldsymbol{q}^{t}_{t}]$.

        \FOR{t=$1,2,\ldots,T$}
            \IF{$t>1$}
                \STATE Stage1: \textbf{Experience-Based Learning}, discovering new task related experiences in already completed tasks. $${\boldsymbol{q}_{-t}},{\boldsymbol{K}_{-t}}  =  \argmin_{{\boldsymbol{q}'_{-t}},{\boldsymbol{K}'_{-t}}}  \mathcal{L}_t  (\medmath{\boldsymbol{q}'_{-t},{\boldsymbol{K}'_{-t}}  \mid \mathcal{D}_{t} ; \mathcal{M}_0, {\boldsymbol{\Phi}_{1}}, \ldots, {\boldsymbol{\Phi}_{t-1}}})$$
            \ENDIF
            \STATE Stage2: \textbf{Novel Knowledge Expansion}, compensates for new knowledge beyond old tasks knowledge. $${\boldsymbol{\Phi}_{t}},{\boldsymbol{q}^{t}_{t}},{\boldsymbol{k}_{t}} =  \mathop{\arg\min} \limits_{{\boldsymbol{\Phi}'_{t}},{\bar{\mathbf{q}}^{t}_{t}},{\boldsymbol{k}'_{t}}}  \mathcal{L}_t ({\boldsymbol{\Phi}'_{t}},{\bar{\boldsymbol{q}}^{t}_{t}},\boldsymbol{k}'_{t} \mid \mathcal{D}_{t} ; \mathcal{M}_0,  \nonumber {\boldsymbol{\Phi}_{1}}, \ldots, {\boldsymbol{\Phi}_{t-1}}, {\boldsymbol{q}_{-t}},{\boldsymbol{K}_{-t}} )$$
            
        \ENDFOR
    \end{algorithmic}
\end{algorithm}

\subsection{Evaluation of Forgetting}
In our experiments, for each task when we continue to expand the knowledge for subsequent tasks, there will exist some extra adapter modules after the novel expansion knowledge. We can discard the extra adapter modules for each task, which is the same as CLiMB~\cite{srinivasan2022climb}. In this case, the forgetting is zero because we do not use the knowledge of the additional adapter after the corresponding step. To implement this in our method, we can simply set the task-specific query of the additional adapters to zero vector, which produces the knowledge coefficient for the corresponding adapter as zero.
However, this implementation comes with a price, how do the subsequent adapters continue to improve the performance of the current task? To overcome this, we also investigate another solution by continue training the task-specific query in the upstream $t$-th step model on old task $k$ ($k \in \{1,\ldots,t-1\}$), which will create a testbed for evaluating the influence of the additional adapter. The optimization is as follows:
\begin{align}
{\boldsymbol{q}_{k}}  =  \argmin_{\boldsymbol{q}'_{k}} \mathcal{L}_k 
 (\medmath{\boldsymbol{q}'_{k}  \mid \mathcal{D}_{k} ; \mathcal{M}_0,  
     {\boldsymbol{\Phi}_{1}}, \ldots, {\boldsymbol{\Phi}_{t}}}).
\end{align}

\section{Ablation Study}
\subsection{Ablation Study for Different Task Order}

We conducted experiments with different upstream task orders, i.e., Uni-First (PIQA$\rightarrow$SNLI-VE$\rightarrow$VQAv2$\rightarrow$iNaturalist2019$\rightarrow$SST-2), to validate that the effectiveness of our method is unrelated to the task order.
Table~\ref{tab:ablation} shows the comparison of accuracy and forgetting with different task orders.
We discovered that the performance achieved after each task's novel knowledge expansion is quite similar. Different sequences contain different amounts of adapter knowledge in experience-based learning, so there are some subtle differences in performance.
For instance, $\textit{Uni-First}$ has added the SNLI-VE adapter knowledge than $\textit{Multi-First}$, and accordingly, the performance has somewhat improved.
In addition, the $\textit{Forgetting}$ results show that for different task sequences, our method is a continuous learner without forgetting.

We conduct experiments to ablate the computation of knowledge coefficients, testing various combinations of adapter-level weighting and token-level weighting. We find that the performance of adapter-level weighting is the same as that of token-level weighting, while adapter-level weighting has fewer learnable parameters. Therefore, we use adapter-level weighting in our methods. 
\begin{table}[t]
    \centering
    \setlength{\tabcolsep}{5pt}
    \resizebox{0.9\linewidth}{!}{%
    \begin{tabular}{lccccc}
    \toprule
    {Multi-First}  & {SNLI-VE \footnotesize{(t=1)}} & {PIQA \footnotesize{(t=2)}} & {iNat2019 \footnotesize{(t=3)}} & {VQAv2 \footnotesize{(t=4)}} &{SST-2 \footnotesize{(t=5)}} \\
    \midrule
    {$\text{Know. Exp}\text{ (Acc.)}$} & 75.15 & 59.88 & 63.30 & 64.73 & 82.48 \\
    {$\text{Forgetting}$} & 0.38 & 0.22 & -0.47 & 0.92 & 0.0 \\
    \midrule
    {Uni-First} & {PIQA \footnotesize{(t=1)}} & {SNLI-VE \footnotesize{(t=2)}}  & {VQAv2 \footnotesize{(t=3)}} & {iNat2019 \footnotesize{(t=4)}} & {SST-2 \footnotesize{(t=5)}}\\
    \midrule
    {$\text{Know. Exp}\text{ (Acc.)}$} & 59.10 & 75.25 & 64.89 & 64.16 & 81.94 \\
    {$\text{Forgetting}$} & -0.57 & 0.26 & 0.29 & 0.29 & 0.0 \\
    \bottomrule
    \end{tabular}}
    \caption{Comparison of accuracy and forgetting with different orders of upstream tasks sequential. $\small\textit{Know. Exp}{\textit{ (Acc.)}}$ represents the accuracy after the stage of novel knowledge expansion on each task, while $\small \textit{Forgetting}$ is the difference of the average accuracy of each time step model in previous-seen tasks to the accuracy after the novel knowledge expansion stage.}
    \label{tab:ablation}
\end{table}

\subsection{Ablation for Knowledge Vector Learning}

We devise a knowledge vector learning method that incorporates the knowledge from different adapters, where each adapter's output contains a set of tokens with the same dimensions. To explore effective methods for fusing different adapter knowledge, we conducted experiments with different linear combination methods, including adapter-level fusion and token-level fusion. The adapter-level fusion uses one knowledge coefficient value for all tokens in an adapter, while each token has a specific coefficient value at the token-level fusion.

\begin{table}[t]
    \centering
    \setlength{\tabcolsep}{1pt}
    \resizebox{0.7\linewidth}{!}{%
    \begin{tabular}{lcccc}
    \toprule
    {Multi-First}  & {SNLI-VE \footnotesize{(t=1)}} & {PIQA \footnotesize{(t=2)}} & {iNat2019 \footnotesize{(t=3)}} & {VQAv2 \footnotesize{(t=4)}} \\
    \midrule
    {token-level} & 74.39 & 59.28 & 62.96 & 64.49 \\ 
    {adapter-level} & 75.15 & 59.88 & 63.30 & 64.73 \\
    \bottomrule
    \end{tabular}}
    \caption{Comparison of the various combinations of computing of knowledge coefficients. The $\textit{token-level}$ represents the linear combination of various adapters at the token level. The $\textit{adapter-level}$ uses the global linear weighting to fuse various adapters.}
    \label{tab:suppl_table1}
\end{table}

Table~\ref{tab:suppl_table1} shows the results of the comparison of different linear combination methods. 
We observe that adapter-level fusion consistently outperforms token-level fusion in all cases. We speculate that there is consistency among tokens in the adapter output, and using a global knowledge coefficient for all tokens usually produces decent results. 
In addition, the token-level fusion introduces more learnable parameters for learning the knowledge coefficient. Specifically, an adapter that uses token-level fusion requires $441.6$k learnable parameters for the knowledge coefficient, while the adapter itself has a parameter size of $883.2$k. This makes the token-level fusion is less parameter-efficient. However, at the adapter-level fusion, only $18.4$k parameters are involved in learning knowledge coefficients, which is negligible compared to the $883.2$k parameters required for an adapter. Moreover, involving more parameters can make optimization difficult, and it becomes challenging to learn the optimal knowledge coefficients.

In our experiments, we utilize adapter-level fusion to calculate the knowledge coefficient, which is not only parameter-efficient but also effectively combines different adapter knowledge.

\section{Additional Results}

\begin{figure*}[t]
  \centering
   \includegraphics[width=1.0\textwidth]{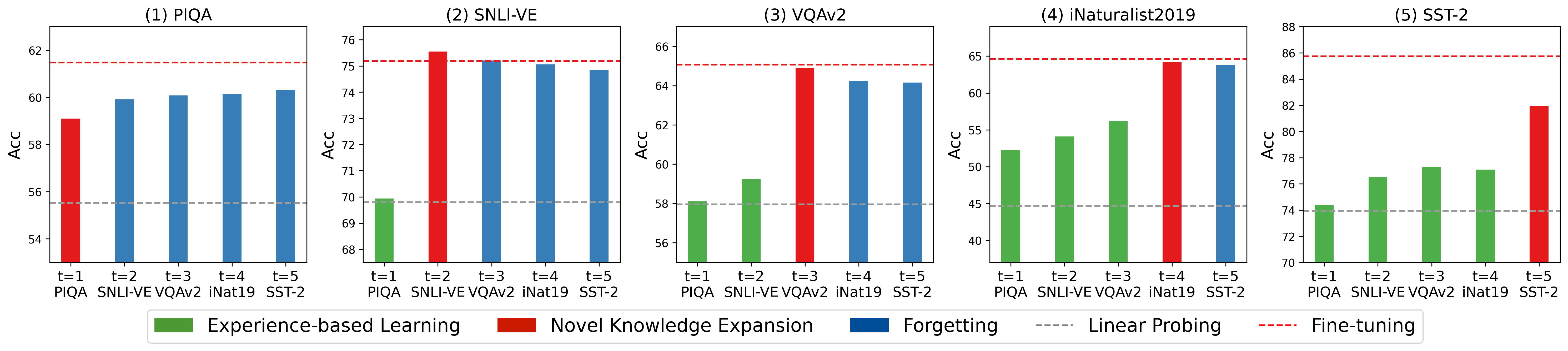}
   \caption{Learning process of upstream continuous learning with Uni-First order (PIQA$\rightarrow$SNLI-VE$\rightarrow$VQAv2$\rightarrow$iNaturalist2019$\rightarrow$SST-2). {Linear Probing} is the lower bound of continual learning while {Fine-tuning} serves as the upper bound. We estimate experience-based learning before novel knowledge expansion in each task, and estimate forgetting afterward.}
   \label{fig:suppl_result-figure1}
\end{figure*}

\begin{figure*}[t]
  \centering
   \includegraphics[width=1.0\textwidth]{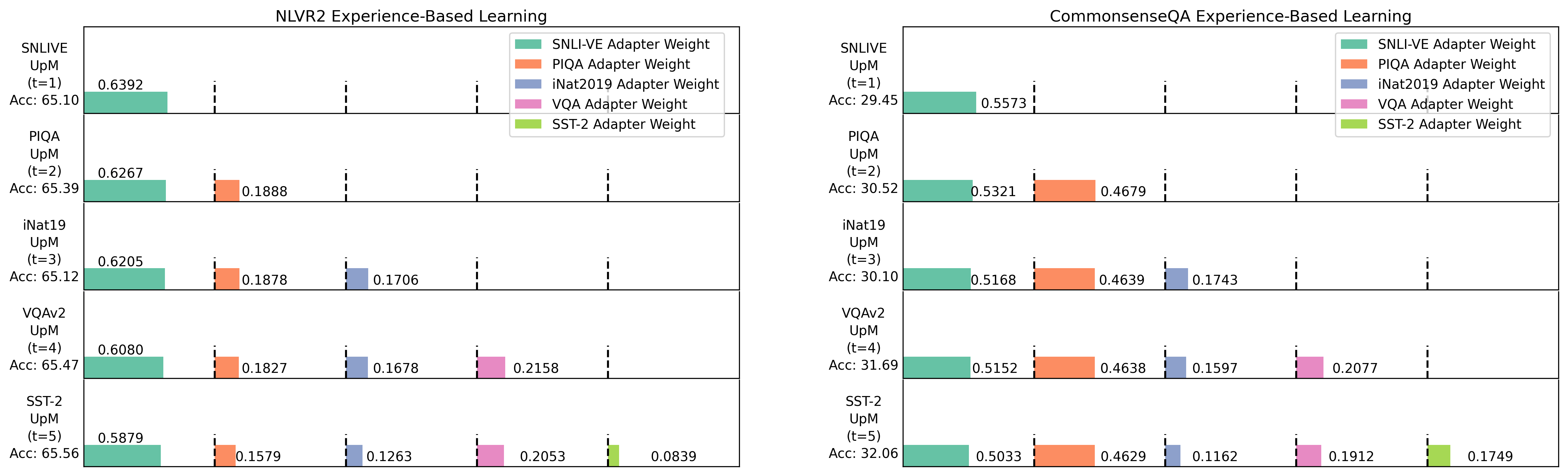}
   \caption{Accuracy and knowledge coefficient comparison of experience-based learning at each step of upstream continuous learning on the NLVR2 and CommonsenseQA datasets. The length of each bar is a knowledge coefficient corresponding to each adapter.}
   \label{fig:suppl_experience-learning}
\end{figure*}

\subsection{Upstream Continual Learning with Uni-First Upstream Task Order}

Fig.~\ref{fig:suppl_result-figure1} shows that the process of upstream continual learning with various task sequential, Uni-First (PIQA$\rightarrow$SNLI-VE$\rightarrow$VQAv2$\rightarrow$iNaturalist2019$\rightarrow$SST-2), which involves training uni-modal tasks first and immediately following multi-modal tasks. Before the stage of novel knowledge expansion in each task, we access the ability to learn from experience-based learning and evaluate forgetting afterward.
Similarly, we introduce $\textit{Linear probing}$ and $\textit{Fine-tuning}$ as references, representing our lower and upper bounds.

Despite using different task orders, we still achieved performance close to fine-tuning on each task under the Uni-First task sequential case. Upon observing forgetting, we find that the utilization of subsequent task knowledge almost did not change the performance on each task, except for a subtle improvement on the PIQA dataset. The constant improvement in experience-based learning reflects that incremental knowledge can assist in decomposing a new task over previously saved knowledge. This indicates that our two-stage learning paradigm is excellent for continual learners without forgetting.

\subsection{Additional Adapter Weighting Analysis}

To further explore the influence of expanding multi-modal and uni-modal task knowledge on downstream tasks, we calculate the knowledge coefficient of the adapter during downstream experience-based learning on the VCR and CommonsenseQA datasets. Fig.~\ref{fig:suppl_experience-learning} is the comparison of accuracy and knowledge coefficient at each step of upstream continual learning on VCR and CommonsenseQA datasets.
We find that the performance improvement on each task is attributed to the utilization of novel incremental knowledge. For instance, the accuracy of the NLVR2 dataset increased by $0.25$ from $t=2$ to $t=3$ as a result of incorporating knowledge from the VQAv2 dataset.
For multi-modal tasks, the expansion of uni-modal tasks has little effect on performance improvement and even has a negative impact in some cases, while the expansion of multi-modal task knowledge usually improves performance.
For uni-modal tasks, knowledge expansion from uni-modal tasks with the same modality or multi-modal tasks usually improves performance, while uni-modal tasks with different modalities often do not improve or have a negative impact that may be attributed to competition between tasks.
In addition, we have observed that as novel knowledge constantly expands, the coefficient of previous knowledge decreases, which means that the decomposition of new tasks over previously saved knowledge decreases when expanding novel knowledge.
We speculate that this may be attributed to the misalignment and competition among different task knowledge, and we will study this in our future work.

\end{document}